\documentclass[pdflatex,sn-mathphys-num]{sn-jnl} 

\usepackage{hyperref}

\usepackage{placeins} 
\usepackage{xcolor}

\newcommand\blfootnote[1]{%
  \begingroup
  \renewcommand\thefootnote{}
  \footnote{#1}%
  \addtocounter{footnote}{-1}%
  \endgroup
}

\usepackage{graphicx}%
\usepackage{multirow}%
\usepackage{amsmath,amssymb,amsfonts}%
\usepackage{amsthm}%
\usepackage{mathrsfs}%
\usepackage[title]{appendix}%
\usepackage{xcolor}%
\usepackage{textcomp}%
\usepackage{manyfoot}%
\usepackage{booktabs}%
\usepackage{algorithm}%
\usepackage{algorithmicx}%
\usepackage{algpseudocode}%
\usepackage{listings}%
\usepackage{tabularx,booktabs,multirow,array}
\usepackage{array}                
\usepackage[table]{xcolor}        
\usepackage{booktabs}             

\theoremstyle{thmstyleone}%
\theoremstyle{thmstyletwo}%

\theoremstyle{thmstylethree}%

\begin{document}

\title[Article Title]{Article Title}


\title[Article Title]{Heuristic Style Transfer for Real-Time, Efficient Weather Attribute Detection
}

\author*[1]{\fnm{Hamed} \sur{Ouattara}}\email{Ouattarahamed6639@gmail.com}
\author[1]{\fnm{Pierre} \sur{Duthon}}\email{pierre.duthon@cerema.fr}
\equalcont{These authors contributed equally to this work.}
\author[3]{\fnm{Pascal Houssam} \sur{Salmane}}\email{pascal.salmane@cerema.fr}
\equalcont{These authors contributed equally to this work.}
\author[1]{\fnm{Frédéric} \sur{Bernardin}}\email{frederic.bernardin@cerema.fr}
\equalcont{These authors contributed equally to this work.}
\author[2]{\fnm{Omar} \sur{Ait Aider}}\email{omar.ait-aider@uca.fr}
\equalcont{These authors contributed equally to this work.}

\affil*[1]{\orgdiv{Intelligent Transportation Systems Research Team(I.T.S)}, \orgname{Cerema}, \orgaddress{\street{8, rue Bernard Palissy}, \city{Clermont-Ferrand }, \postcode{63017},  \country{France}}}
\affil[2]{\orgdiv{Clermont Auvergne INP}, \orgname{Université Clermont Auvergne, Pascal Institute, CNRS}, \orgaddress{\street{4 Av. Blaise Pascal}, \city{Clermont-Ferrand}, \postcode{63178}, \country{France}}}
\affil*[3]{\orgdiv{Intelligent Transportation Systems Research Team(I.T.S)}, \orgname{Cerema}, \orgaddress{\street{ 1 avenue du Colonel Roche}, \city{Toulouse }, \postcode{31400},  \country{France}}}

\abstract{
We present lightweight and efficient architectures to detect weather conditions from RGB images, predicting the weather type (sunny, rain, snow, fog) and 11 complementary attributes such as intensity, visibility, and ground condition. With a total of 53 classes distributed among the tasks. This work examines to what extent weather conditions manifest as variations in visual style. We investigate style-inspired techniques—Gram matrices, a truncated ResNet-50 (targeting lower/intermediate layers), and PatchGAN-style architectures—within a multi-task framework with attention mechanisms. Two families are introduced: RTM (ResNet50-Truncated-MultiTasks) and PMG (PatchGAN-MultiTasks-Gram), along with their variants. Contributions include: automation of Gram-matrix computation, integration of PatchGAN in supervised multi-task learning, and local style capture via local Gram for improved spatial coherence. We also release a dataset of 503\,875 images annotated with 12 weather attributes, under a Creative Commons Attribution (CC-BY) license. The models achieve F1 scores above 96\,\% on our internal test set and beyond 78\,\% in zero-shot evaluation on several external datasets, confirming their generalization ability. The PMG architecture, with fewer than 5 million parameters, runs in real time with a small memory footprint, making it suitable for embedded systems. The overall modularity of the models allows tasks related to style or weather to be added or removed as needed.

}


\maketitle
\blfootnote{\textit{Funded by the European Union (grant no. 101069576). However, the opinions and points of view expressed are solely those of the author(s).} Hamed Ouattara, Pierre Duthon, Frédéric Bernardin, and Pascal Salmane are with Cerema (I.T.S Team, France). Omar Ait Aider is with Institut Pascal, France.}

\section{Introduction}\label{sec1}
\FloatBarrier
\begin{figure}[H]
    \centering
    \includegraphics[width=0.9\columnwidth]{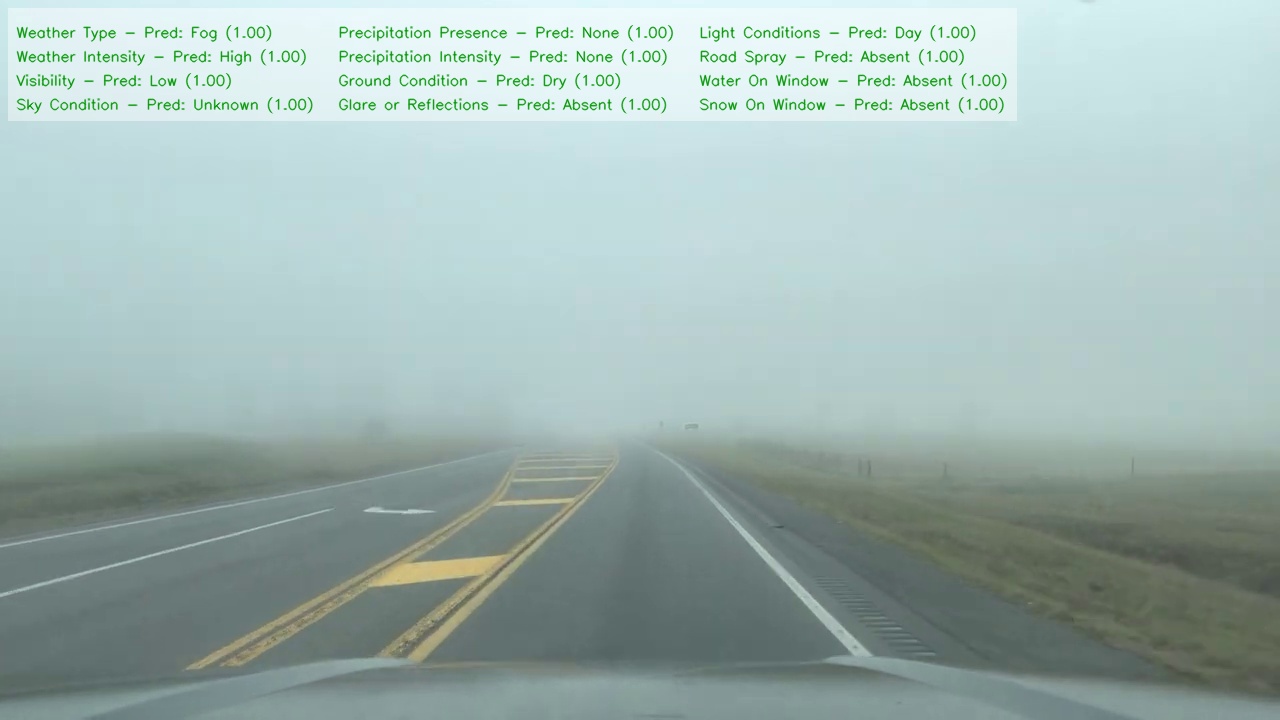}
    \caption{Joint detection of 12 weather-related attributes with ResNet50-Truncated-MultiTasks (RTM). 
This example illustrates the simultaneous prediction of Weather Type, Visibility, etc. The aim is to describe the environment from the standpoint of weather and its effects, using 12 key characteristics (e.g., Ground Condition, useful for adapting speed on wet pavement) with a single real-time model. 
The architecture is modular: task heads can be disabled at inference to reduce computational cost, without impacting other tasks thanks to independent classifiers and attention mechanisms.}
    \label{fig:figure0}
\end{figure}

\FloatBarrier

Models capable of detecting in real time weather conditions and their effects (\textit{ground state — wet, dry, snowy —, visibility, precipitation, glare})(\hyperref[fig:figure0]{Fig.~\ref{fig:figure0}}) from RGB images captured by day and by night would open up new prospects for security, transport, energy, and climate.

In transport, such a system would optimize autonomous driving and ADAS \cite{engstle2023adverse,shahzad2024robustadasenhancingrobustness,8666747} by dynamically adapting speed, alerts, and signage to actual conditions (\textit{rain, snow, fog}). Integrated into traffic management, it would help smooth traffic flow and reduce accident risk.

In energy and climate, automated detection could facilitate regulation of renewable energy production (\textit{solar, wind}) according to local weather, while improving the energy efficiency of infrastructure (\textit{lighting, heating, signage}). Leveraging widely deployed road cameras would further provide continuous ultra-local monitoring of conditions, strengthening the accuracy of weather forecasts.

Finally, thanks to their lightweight nature and speed, such models could be embedded in devices like dashcams or drones, offering a practical, low-power, and easily deployable solution.

Despite these promises, visual weather detection remains challenging. As detailed in Section~\ref{sec:review}, many current approaches are limited to recognizing the weather type (\textit{sunny, rain, snow, fog}) without capturing secondary effects, or they rely on computational resources that are too heavy for real-time execution.

\begin{figure}[!t]
    \centering
    \includegraphics[width=0.9\columnwidth]{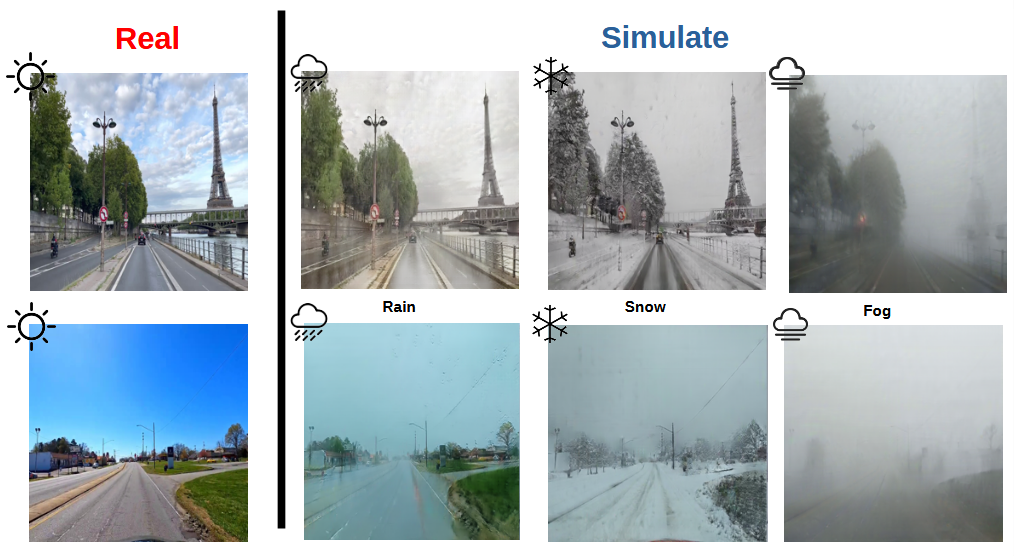}
    \caption{Simulation of weather conditions via style transfer with CycleGAN~\cite{key2}. Left columns: real (source) images. Right columns: stylized outputs reproducing rain, snow, or fog at varying intensities. This example shows that an unpaired style-transfer model can simulate visual signatures of weather (rain streaks, luminous haze, color drift), illustrating that weather-related changes can be expressed as changes in visual style and motivating our investigation of this link.}
    \label{fig:figure1}
\end{figure}

In this paper, we address the detection of meteorological characteristics in an RGB image from the perspective of visual style. Our approach stems from the observation that certain style transfer algorithms, such as CycleGAN~\cite{key2} or CUT~\cite{key3}, can simulate different weather conditions, as illustrated in \hyperref[fig:figure1]{Fig.~\ref{fig:figure1}}. This suggests an exploitable relationship between weather conditions and visual style.

We hypothesize that weather conditions often affect the appearance of objects in a scene in a way that can be interpreted as a change in visual “style”, and we investigate how far this intuition can be pushed in practice. In a sunny, clear scene, objects receive a broad portion of the solar light spectrum, revealing vivid, high-contrast colors.
Conversely, during a storm under an overcast sky, the light is heavily filtered, which darkens the scene and softens the hues.  Wet surfaces become more reflective (with water acting as a mirror), fog diffuses contours and reduces visibility, while snow homogenizes the scene under a light veil. Imagine the same scene observed under these different weather conditions (\hyperref[fig:figure1]{Fig.~\ref{fig:figure1}}): it becomes clear that they primarily alter the visual appearance of objects rather than their intrinsic nature. Much like a painter imprinting a personal touch on each work, every weather condition confers a distinctive stylistic signature on the scene. Moreover, several early works already suggest that weather can be inferred from an image’s overall appearance. Methods based on hand-crafted descriptors of brightness, contrast, sharpness, saturation, sky structure, reflections, or haze~\cite{ship2024real, zhao2011feature, lu2014two, roser2008classification} show that it is possible to distinguish different weather types or estimate meteorological variables from simple illumination and texture cues. Although low-dimensional and non-deep, these approaches support the idea that weather conditions primarily manifest as appearance signatures—akin to what we refer to here as visual \textit{style}.

Our goal is not to formally prove that weather modifies the visual style of an image, but to start from an empirical observation: some style transfer algorithms can realistically reproduce different weather conditions. This behaviour suggests the existence of an exploitable link between meteorological characteristics and the visual style of a scene.

In this work, we exploit this link to design models capable of automatically detecting weather conditions from images. The central question we ask is the following: \emph{are the stylistic properties of an image, by themselves, sufficiently informative to characterize its meteorological state?} Our experiments do not provide a complete answer to this question, but the results show that a family of style-sensitive architectures already constitutes an effective tool for predicting several weather-related attributes.

More precisely, we seek to understand why certain architectures learn representations that are sensitive to visual style, and how to use them to design models capable of identifying not only the weather type (\textit{sunny, snow, fog, rain}), but also more specific attributes, as illustrated in \hyperref[fig:figure0]{Fig.~\ref{fig:figure0}}.

\begin{figure}[!t]
    \centering
    \includegraphics[width=0.8\columnwidth]{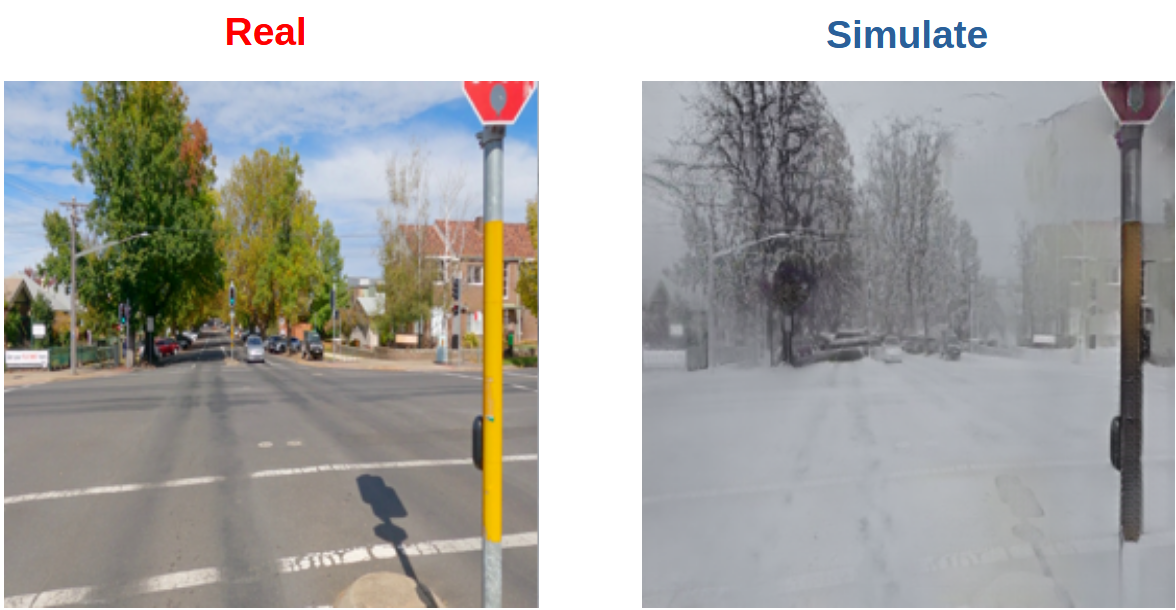}
    \caption{Snow simulation via style transfer with CycleGAN~\cite{key2}. Left: real image; right: “snow” stylized image reproducing coherent cues (overcast sky, snow-covered ground, attenuated vegetation, tracks). These results suggest that unpaired style transfer, guided by a PatchGAN discriminator, can capture stylistic signatures related to multiple meteorological attributes—which motivates our multi-task approach, where the local sensitivity of PatchGAN is treated as a potential “sensor” for such cues.
}
    \label{fig:figure2}
\end{figure}

This sensitivity to appearance and style is one of the motivations for our approach: as shown in Fig.~\ref{fig:figure2}, models such as CycleGAN can reproduce realistic visual details—for example, by removing leaves from trees to simulate winter conditions, or by adding footprints on worn asphalt—which suggests that style-sensitive mechanisms may be useful for weather-related prediction.

Concretely, we introduce two main architectures: \textbf{RTM} (ResNet50-Truncated-MultiTasks) and \textbf{PMG} (\hyperref[sub-sec:PatchGan]{PatchGAN}-MultiTasks-\hyperref[sub-sec:gram]{Gram}), along with their variants. These models, deliberately biased toward style-sensitive features, learn to extract, weight, and classify appearance cues that we hypothesize to be informative for weather-related attributes.

Our architectures detect not only the weather type, but also 11 secondary effects related to weather conditions with 53 classes   (\hyperref[fig:figure0]{Fig.~\ref{fig:figure0})}:
\begin{itemize}
    \item Weather type (sunny, rain, snow, fog),
    \item Intensity (low, medium, high),
    \item Visibility (good, medium, poor),
    \item Sky condition (clear, cloudy, partly cloudy),
    \item Ground condition (dry, wet, snowy),
    \item Presence of precipitation,
    \item Precipitation intensity (low, medium, high),
    \item Presence of glare,
    \item Light conditions (day, night),
    \item Road spray,
    \item Presence of water or snow on the windshield.
\end{itemize}

These architectures are designed to be modular and lightweight. Each task is associated with an attention mechanism and a dedicated classifier. For example, the \textbf{PMG} model has $\approx 2,43\,\mathrm{M}$ parameters, of which (\emph{43\,k} are dedicated to style extraction. Adding a task incurs only a marginal cost of 33{,}285 to 66{,}057 parameters, enabling flexible adaptation to application needs. The source code is available online\footnote{\url{https://github.com/Hamedkiri/Heuristic_Style_Transfer_for_Real-Time_Efficient_Weather_Attribute_Detection}}.  

\section{Literature Review}
\label{sec:review}

To the best of our knowledge, no prior work simultaneously covers the 12 meteorological attributes considered here. This multi-attribute definition stems from the requirements of the European project \href{https://roadview-project.eu/}{ROADVIEW} and aligns with European standards and recommendations \cite{engstle2023adverse, roh2020analysis}: visibility (linked to ADAS stopping distances, Euro-NCAP §8), spray and water on windshield (camera/LiDAR sensor obstruction), glare (risk of sun-glare crashes), and other attributes compatible with WMO metadata.

Large multimodal models (e.g., GPT-4o \cite{bang2023multitaskmultilingualmultimodalevaluation, Gill_2023, liu2023textunveilingmultimodalproficiency}) show broad detection and explainability capabilities but require data volumes and compute resources that are incompatible with real-time embedded execution. We target specialized architectures that are frugal in both data and computation. In what follows, we review (i) early works that support the link between visual “style” and weather, and (ii) the literature specific to each task considered.

\subsection{Approaches Based on Handcrafted Appearance Descriptors}
\label{subsec:handcrafted-style-like}

Several prior studies implicitly assume that weather conditions manifest first as appearance signatures—contrast, brightness, sky hue, haze, or reflections—before being described in terms of objects.

Roser and Moosmann~\cite{roser2008classification} offer one of the earliest approaches using a single onboard color camera. Their system extracts color histograms, contrast, and texture measures, then feeds a SVM to distinguish situations (\emph{sunny}, \emph{light rain}, \emph{heavy rain}, \emph{fog}, etc.), interpreting global variations in illumination and contrast as weather cues.

Zhao et al.~\cite{zhao2011feature} generalize this idea by explicitly defining spatial and spatiotemporal descriptors for rain, snow, or fog: inter-frame intensity fluctuations, gradients, chrominance, and noise levels are combined in a two-stage classifier. Again, weather is modeled through the dynamics of luminance, contrast, and color rather than through specific objects.

Lu et al.~\cite{lu2014two} focus on a more targeted problem—\emph{sunny} vs \emph{cloudy} from a single image—while pushing the appearance-cue logic further. They construct a “\emph{weather feature}” aggregating cues such as sky color and structure, shadow sharpness, atmospheric veil, reflections, and global contrast, then train a \emph{collaborative learning} framework that accounts for the presence/absence of these cues per image.

More recently, Ship et al.~\cite{ship2024real} return to a frugal, real-time approach by combining a small set of interpretable descriptors (mean brightness, saturation, blur and noise indices, contrast and haze measures) with a SVM to classify surveillance-camera images into a few weather categories (\emph{clear}, \emph{haze}, \emph{rain}, etc.).

Taken together, these works can be viewed as precursors to our approach: they already formalize, via handcrafted descriptors, a link between meteorological conditions and \emph{visual style} (contrast, brightness, hue, haze), at the cost of manual feature engineering. Our contribution is to reinstate this intuition in a unified framework where such stylistic information is learned automatically by truncated-ResNet or PatchGAN architectures, through Gram matrices, implicit patches, and multi-task attention.

\subsection{Weather classification, visibility, illumination}

Weather-condition classification has been extensively studied with CNNs \cite{xia2020resnet15, naufal2022weather, papadimitriou2023advancing, patel2021weather, mittal2023classifying, kukreja2023multi, ibrahim2019weathernet, guerra2018weather, dahmane2020analyse} and, more recently, attention mechanisms \cite{introvigne2024real, li2023study, rani2023weather, pikun2022image}. 
Mittal and Sangwan \cite{mittal2023classifying} explore scaling (transfer learning + Spark) and report 97.77\% accuracy (InceptionV3+LR) on a small Kaggle corpus (1{,}125 images). Kukreja \textit{et al.}\ \cite{kukreja2023multi} combine CNN and SVM. Rani \textit{et al.}\ \cite{rani2023weather} couple convolutional self-attention (CSA) with dilated spatial attention (ASA) and reach 97.47\% on MWD. These approaches are effective but rarely tailored to real-time embedded settings.

Truncating deep networks improves the performance/compute trade-off. Xia \cite{xia2020resnet15} truncates ResNet-50 (ResNet15) and raises F1 from 85.76\% to 96.03\% on WeatherDataset-4 (4{,}983 images, 4 classes). Mahesh \cite{dhananjaya2021weather} uses ResNet-18 for three weather types and three illumination levels, with a similar gain. These works do not deeply analyze the mechanisms behind the improvements; our framework interprets these results through style cues (see Section~\ref{sec:background}) and leverages them in RTM/RTMG to predict, beyond weather type, 11 additional attributes.

For visibility, several studies rely on CNNs, sometimes guided by physical models \cite{you2021dmrvisnetdeepmultiheadregression, Ferreira_Neto_2022, chaabani2018estimating}; \cite{you2021dmrvisnetdeepmultiheadregression} proposes a dedicated multi-head regression.

\subsection{Ground condition}

Characterizing ground condition (dry, wet, snowy) is critical for safety \cite{basavaraju2019machine, piroli2024semanticspray++, sattar2018road, jagatheesaperumal2023artificial, gragnaniello2025real, chen2016inexpensive, almalioglu2022deep, dilorenzo2023use, changalvala2019lidar, panhuber2016recognition}. 
Piroli \textit{et al.}\ \cite{piroli2024semanticspray++} release SemanticSpray++, a multimodal dataset (camera/LiDAR/radar) focused on wet pavement. Jagatheesaperumal \textit{et al.}\ \cite{jagatheesaperumal2023artificial} exploit onboard acoustic signals (Mel spectrograms) to classify road surfaces (including \emph{slippery}), reporting up to 98\% with an MLP on custom and public sets. These pipelines benefit from specialized sensors; our goal is reliable estimation from a single RGB camera, to remain compatible with cost/energy constraints.

\subsection{Weather intensity and water/snow on glass}

Fiallos-Salguero \cite{manuel2025toward} proposes a hybrid framework to estimate urban rainfall from surveillance videos, correlated with rain gauges (R\textsuperscript{2} from 0.89 to 0.93) and robust day/night. Zheng \cite{zheng2023toward} develops a two-stage approach for real-time rain intensity (CCTV, 120\,h; R\textsuperscript{2}\,=\,0.92; MAE\,=\,0.89\,mm/h; 4.8\,fps). Carvalho \cite{carvalho2023modelling} combines a physical model and dimensional analysis to predict perceived intensity on a moving vehicle, validated in a wind tunnel and real conditions, with strong correlation to measurements. Access to datasets with rain gauges remains limited, hindering replication and systematic comparison.

\paragraph{Summary.}
The literature shows: (i) marked progress via CNN/attention and truncation for weather classification; (ii) frequent reliance on costly sensors (LiDAR, radar, acoustics) for ground condition and intensity; (iii) few real-time embedded solutions; (iv) the absence of integrated frameworks covering many attributes simultaneously. Our contribution targets this space: computation-frugal models, a single RGB camera, and unified multi-attribute coverage. In the next section~\ref{sec:background}, we will explore the methods used in style transfer to capture the style of an image.

\section{Background}
\label{sec:background}

Style transfer methods aim to modify the appearance of an image while preserving its content structure~\cite{key1,key2,key3,gatys2016image}. In practice, however, the boundary between style and content is not sharply defined: reducing illumination mainly alters appearance, but can also make the content partially unobservable. For example, when an object is gradually plunged into darkness, its appearance changes and some parts of its content become increasingly invisible.

More generally, objects are only accessible to us through a certain appearance. Modifying the “style” often amounts to changing the nature of the light with which we observe them. The goal is then not always to preserve exactly the same content, but sometimes to reveal other aspects of it. Thus, when physicists move beyond the visible spectrum (for instance by using infrared imaging), it is precisely to highlight properties of the object that were difficult to perceive in the visible range. All of this illustrates the difficulty of rigorously defining what “style” is from a theoretical point of view, and of drawing a clear separation between style and content.

In this section, we review several style transfer methods proposed in the literature, which handle this style/content distinction in different ways, and we briefly explain how we use them in our approach.

\subsection{Gram matrix and style}
\label{sub-sec:gram}

\begin{figure}[!t]
  \centering
  \includegraphics[width=0.9\columnwidth]{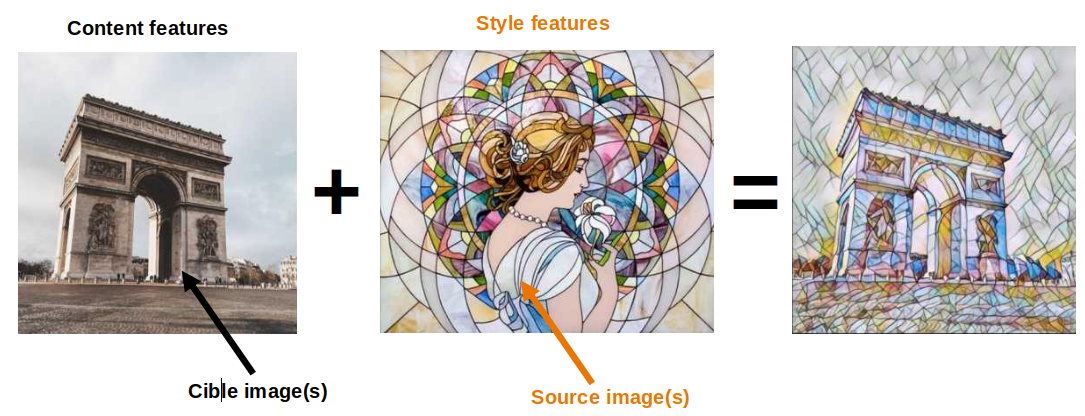}
  \caption{Neural style transfer~\cite{gatys2016image}: a synthetic image is optimized to preserve the structure of the \emph{content} images while reproducing the style statistics (Gram matrices) of the \emph{style} images.}
  \label{fig:figure4}
\end{figure}

In the seminal framework of Gatys et al.~\cite{gatys2016image} (Fig.~\ref{fig:figure4}), the \emph{content} is carried by semantic representations extracted from deep layers of pretrained networks for classification or detection, such as VGG~\cite{simonyan2015deepconvolutionalnetworkslargescale}. Such models learn to group different instances of the same object (e.g., cars with different shapes or colors) into a single class.

The \emph{style}, by contrast, is modeled via correlations between feature maps within a given layer, encoded by a \emph{Gram matrix}. The underlying idea is that style represents appearance regularities in which objects are “immersed”, and which constitute information shared across the entire scene. Let $F\in\mathbb{R}^{C\times H'\times W'}$ be a tensor of feature maps, and $X\in\mathbb{R}^{C\times S}$ its flattened version ($S = H'W'$). Then the Gram matrix
\[
  G \;=\; \tfrac{1}{S}\, X X^\top \in \mathbb{R}^{C\times C}
\]
captures co-activation statistics that are independent of the explicit geometry of objects.

\begin{figure}[!t]
  \centering
  \includegraphics[width=\columnwidth]{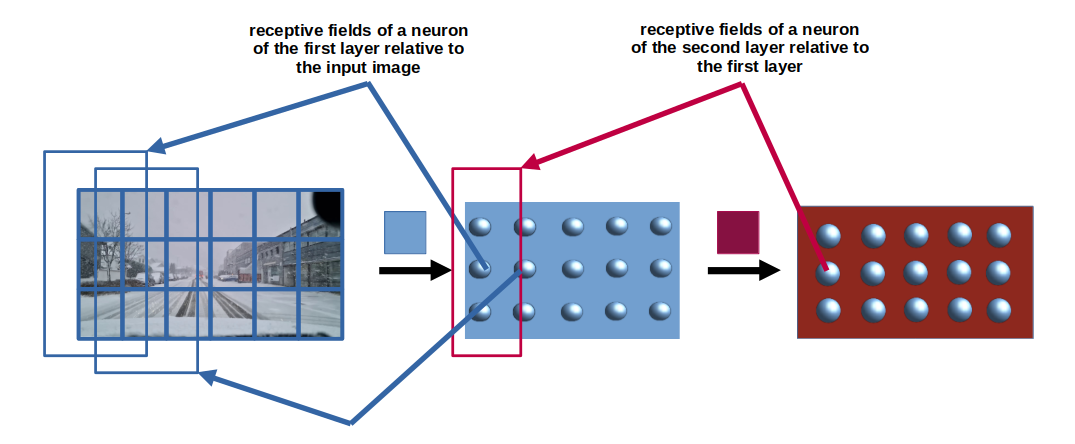}
  \caption{Receptive field in a convolutional network: deeper layers aggregate information over larger spatial extents, and descriptors increasingly reflect global properties.}
  \label{fig:figure5}
\end{figure}

Layer depth modulates the stylistic scope: intermediate layers describe rather local or mesoscale attributes, whereas deeper layers reflect more global properties, in connection with the increase of the \emph{receptive field} (Fig.~\ref{fig:figure5}). This observation sheds light on the gains reported with \emph{truncated} networks (ResNet15/18) compared to a full ResNet50~\cite{xia2020resnet15,dhananjaya2021weather}.

Classical Gram-based style modeling, however, suffers from an important limitation: it discards the spatial arrangement of style (invariance to permutations of positions within a feature map). This can be detrimental when reconstructing localized weather effects (for instance, a denser fog layer in a specific region of the image).

More fundamentally, this way of defining content as a “class concept” relies on a shared intersubjective reality (what we decide to regard as the same object). In a style transfer setting, and even more so in recent diffusion-based approaches~\cite{sohl2015deep,ho2020denoisingdiffusionprobabilisticmodels}, this introduces a potential Achilles’ heel: the algorithm can, in principle, replace a given instance of an object by another instance of the \emph{same} class while still satisfying its definition of content. In safety-critical applications such as driving, this may be problematic (e.g., replacing one traffic sign with another).

In our models, we exploit Gram-based style in two ways:
(i) \emph{RTMG} computes global Gram matrices from a truncated ResNet50 (with \emph{learned} weights, as opposed to the frozen VGG used in~\cite{gatys2016image});
(ii) \emph{PMG} introduces \emph{local} Gram matrices via a PatchGAN-style trunk, in order to preserve the spatial coherence of style.
Both variants are then integrated into a multi-task architecture with attention (Section~\ref{sec:models}).

\subsection{PatchGAN and style}
\label{sub-sec:PatchGan}
Style-transfer algorithms such as Pix2Pix \cite{key1}, CycleGAN \cite{key2}, and CUT \cite{key3} do not propose an explicit separation between style and content. Their operation relies on two mechanisms:
(i) a \emph{content preservation} mechanism, and
(ii) a \emph{style modification} mechanism.

In all three cases, style is modified through adversarial learning~\cite{goodfellow2020generative}. Content is constrained differently: in CycleGAN \cite{key2} by a \emph{cycle-consistency} loss and an L1 penalty; in CUT \cite{key3} by a combination of L1 and patchwise contrastive learning; and in Pix2Pix \cite{key1} by paired images, where L1 or L2 losses suffice.

These approaches, which do not explicitly define the boundary between style and content, operate in a form of tension between maintaining content and transforming style. The balance is automatically determined by the task. For example, in CUT \cite{key3}, transforming cats into dogs implies more content modification than converting horses into zebras or simulating rain/snow (\hyperref[fig:figure1]{Fig.~\ref{fig:figure1}}, \hyperref[fig:figure2]{Fig.~\ref{fig:figure2}}). This flexibility stems from adversarial learning, which dynamically adapts the boundary between style and content to meet varied tasks.

In this setting, the \emph{discriminator}, which guides the generator, adopts in all three architectures a \emph{PatchGAN}-type structure. Introduced by Isola \textit{et al.}~\cite{key1} with Pix2Pix, this discriminator is designed to be sensitive to style. Its specificity lies in the last layer, which perceives the image in fragments through \emph{disjoint receptive fields}: each neuron is influenced by a distinct region of the image, without explicit patch segmentation but through the network’s own perception. This structure enables local evaluation and promotes the capture of \emph{high-frequency details}, which are crucial in style transfer.

PatchGANs cannot be too deep: as explained in the previous subsection, the deeper a network, the larger the neurons’ receptive fields (\hyperref[fig:figure5]{Fig.~\ref{fig:figure5}}). Beyond a certain depth, receptive fields cease to be disjoint and local, reducing the ability to capture fine details.

We exploit this constraint in our \emph{PatchGAN-MultiTasks} and \emph{PatchGAN-MultiTasks-Gram} architectures. By structuring the network so that subgroups of neurons retain disjoint receptive fields, we introduce a novel approach: \hyperref[subsec:pmg-arch]{computing Gram matrices over implicit patches}. Rather than a single matrix for the entire image, we generate multiple Gram matrices, each describing the stylistic correlations of a distinct region. This method, central to \emph{PatchGAN-MultiTasks-Gram}, enables \emph{local style capture}, a new advance in this area.

\subsection{Attention mechanisms}
Originating in NLP~\cite{vaswani2023attentionneed}, attention mechanisms have become standard in vision~\cite{xu2016showattendtellneural, hu2019squeezeandexcitationnetworks, wang2018nonlocalneuralnetworks}. The principle is to weight representations spatially (or by patch/token) according to their relevance for the task: \emph{queries} compare \emph{keys} to produce weights, which are then applied to the \emph{values}.

Our architectures share the same scheme:
(i) a \emph{style-sensitive encoder} (truncated ResNet or PatchGAN trunk) extracts feature maps; 
(ii) for each task, a lightweight single-head attention adaptively aggregates these features, followed by a dedicated classifier.
This factorization allows shared extraction (style) while specializing the decision (task), with limited computational cost.

Attention improves focus on relevant regions, strengthens inter-class separability, and preserves model compactness (low latency, embedded use). Section~\ref{sec:models} details the resulting variants.

\section{Dataset}
\label{sec:data}

To train and evaluate our models, we curated a new dataset of 503{,}875 RGB images, released under a CC-BY license.\footnote{\url{https://github.com/Hamedkiri/Weather_MultiTask_Datasets}}
Resolutions range from 640\,$\times$\,450 to 1{,}280\,$\times$\,720. The images come from 227 public videos under CC0 or Open Licence.

\begin{figure}[!t]
    \centering
    \includegraphics[width=\columnwidth]{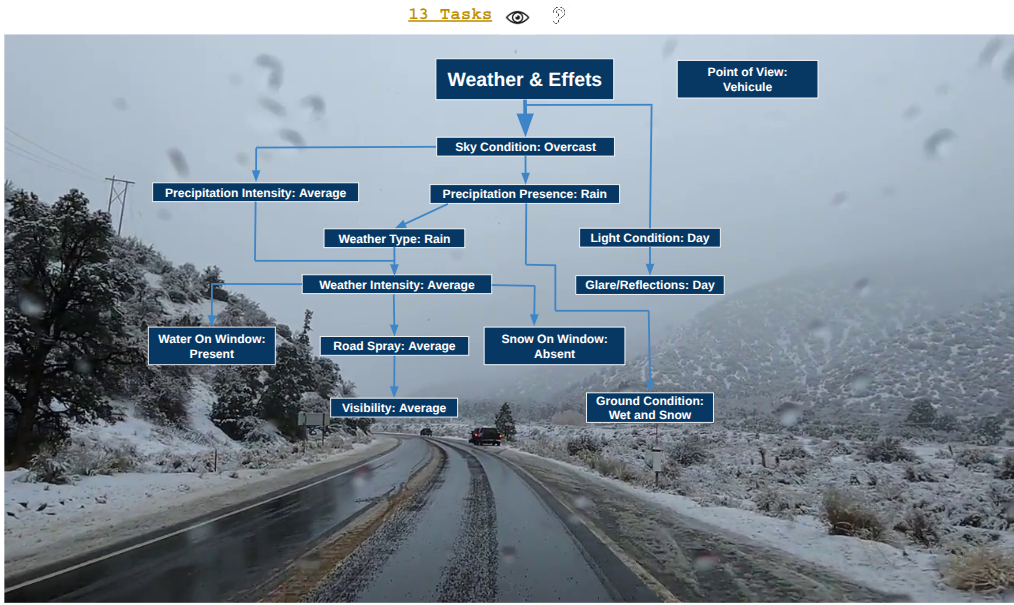}
    \caption{Annotation example. Each image is annotated on 13 criteria(each one has a number of classes, 53 of which are related to the weather). 12 related to weather conditions and 1 related to viewpoint.}
    \label{fig:annot-example}
\end{figure}

\begin{figure}[!t]
    \centering
    \includegraphics[width=\columnwidth]{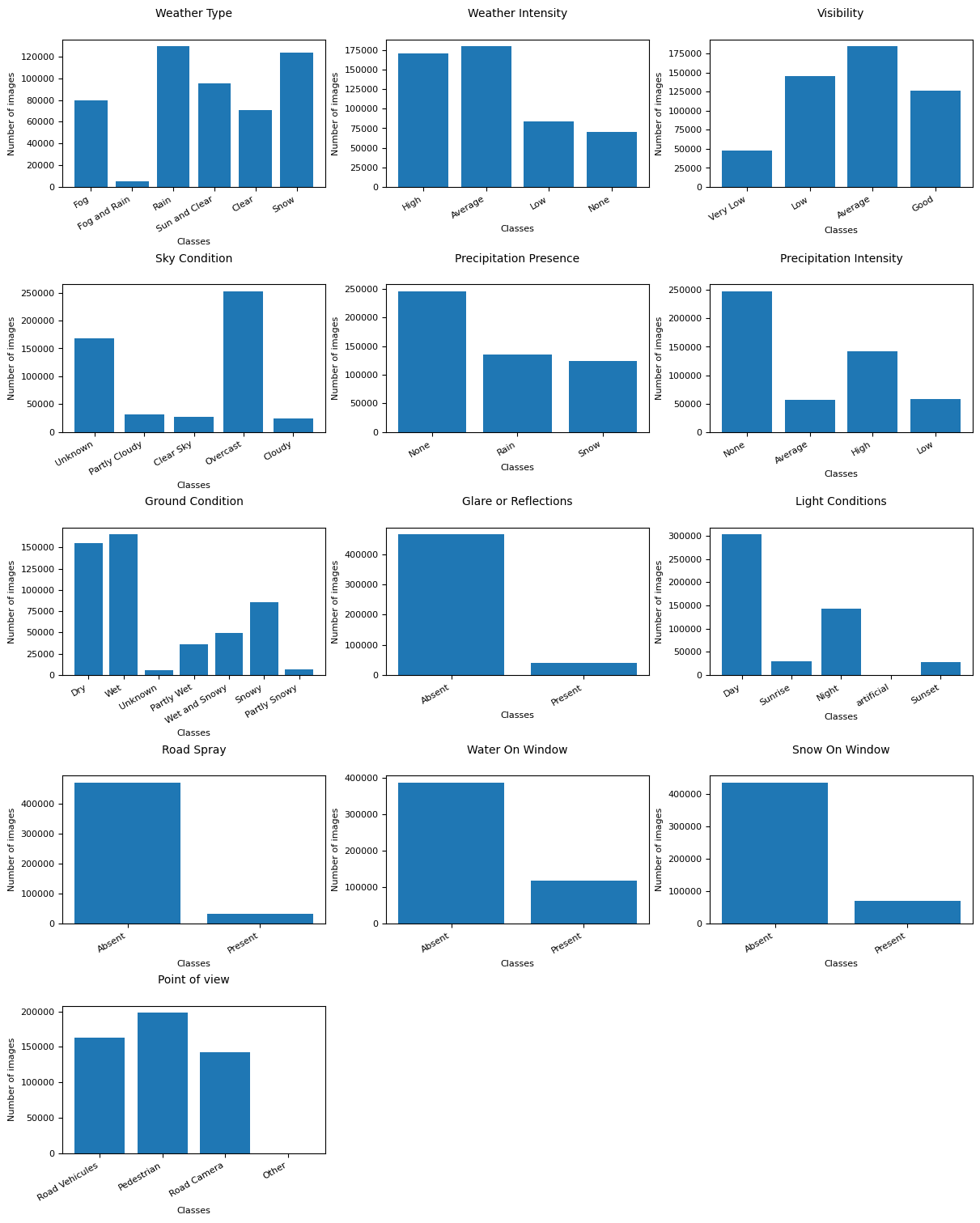}
    \caption{Distribution of the 13(53 classes related to the weather) annotated attributes. The split is near-balanced for Weather Type, but several tasks are strongly imbalanced, notably Road Spray, Water on Windshield, and Snow on Windshield. To limit learning bias, we apply per-task, per-class weighting during training.}
    \label{fig:dist-overview}
\end{figure}

Each image is manually annotated by three independent annotators on 13 criteria (Fig.~\ref{fig:annot-example}), of which 12 concern weather and 1 concerns the \emph{viewpoint} (an informative criterion not used for training, intended to diversify the dataset by perspective). For each image, the 12 attributes(53 classes) are annotated by three annotators. The final label is determined by a majority vote ($\ge$ 2/3). In the absence of a majority (three distinct opinions), the attribute is marked \textit{Unknown} or excluded. This protocol limits the impact of individual disagreements and ensures consistent aggregation.  \hyperref[fig:dist-overview]{Fig.~\ref{fig:dist-overview}} summarizes class distributions across all tasks. The 13 criteria with their 56 classes are:

\begin{itemize}
    \item Weather Type: Clear, Sunny\,+\,Clear, Rain, Snow, Fog, Fog\,+\,Rain, Fog\,+\,Snow, None.
    \item Weather Intensity: Low, Medium, High, None.
    \item Visibility: Very Low, Low, Medium, Good.
    \item Sky Condition: Unknown, Clear Sky, Partly Cloudy, Cloudy, Overcast, Partly Overcast.
    \item Precipitation Presence: None, Rain, Snow, Hail.
    \item Precipitation Intensity: None, Low, Medium, High.
    \item Ground Condition: Dry, Wet, Partly Wet, Snowy, Partly Snowy, Wet\,+\,Snowy, Unknown.
    \item Glare / Reflections: Absent, Present.
    \item Light Conditions: Day, Night, Sunset, Sunrise, Artificial Light.
    \item Water Spray (road spray): Absent, Present.
    \item Water on Windshield: Absent, Present, None.
    \item Snow on Windshield: Absent, Present, None.
    \item Viewpoint: Onboard Vehicle, Pedestrian, Fixed Road Camera.
\end{itemize}

\begin{figure}[!t]
    \centering
    \includegraphics[width=0.9\columnwidth]{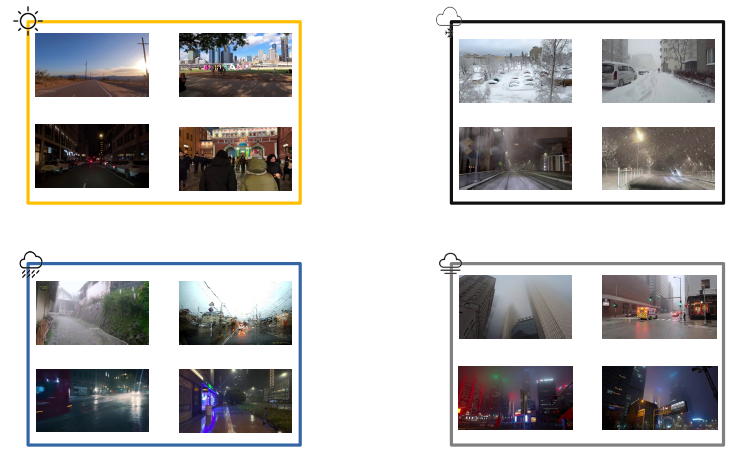}
    \caption{Example images showing rain, snow, fog, and clear weather, by day and by night.}
    \label{fig:dataset-examples}
\end{figure}

\subsection{Construction of training and test subsets}
\label{subsec:data-splits}

We split the corpus (503{,}875 images from 227 videos) into two subsets that are disjoint at the image level, while limiting temporal redundancy and balancing the contribution of each source.

\paragraph{Principles.}
\begin{itemize}
  \item \emph{Source stratification}: all videos contribute to both \emph{train} and \emph{test}; no image is shared.
  \item \emph{Redundancy reduction}: random selection with a spacing constraint between frames within the same video.
  \item \emph{Per-source balancing}: per-video quotas to prevent any source from dominating the distribution.
\end{itemize}

\paragraph{Sizes and effects.}
\begin{itemize}
  \item \emph{Full set}: mean gap = 38.63.
  \item \emph{Train}: 250{,}000 images; mean frame-index gap within a video = 62.75.
  \item \emph{Test}: 25{,}000 images; mean gap = 427.65.
  
\end{itemize}

\paragraph{Reproducibility.}
Lists of image IDs, the seed, and split-generation scripts are provided in \href{https://github.com/Hamedkiri/Weather_MultiTask_Datasets}{\texttt{the GitHub repository}}. Additional details (sampling log, spacing settings, per-source quotas) are also available there.

\section{Methods and Presentation of Our Models.}
\label{sec:models}

We present two model families: (i) variants of \emph{truncated ResNet-50} and (ii) \emph{PatchGAN}-based models. Gatys et al.~\cite{gatys2016image} show that low/intermediate layers of a pretrained network (e.g., VGG~\cite{simonyan2015deepconvolutionalnetworkslargescale}) capture \emph{style} more effectively. In the same vein, several studies~\cite{xia2020resnet15,dhananjaya2021weather} report that truncating ResNet-50 improves performance. We therefore \emph{automate} the choice of the truncation point via evolutionary search (Sec.~\ref{subsec:evo-hpo}) and, unlike prior work, propose a \emph{multi-task} architecture with independent heads.

In addition, we use PatchGAN in a novel way: its disjoint receptive fields are leveraged to compute \emph{local Gram} statistics (per patch), mitigating the loss of spatial stylistic coherence highlighted by~\cite{gatys2016image} when Gram matrices are applied globally.

\subsection{RTM model: truncated ResNet \& per-task attention over spatial tokens}
\label{subsec:mt-rtm-arch}

\begin{figure}[h]
    \centering
    \includegraphics[width=\columnwidth]{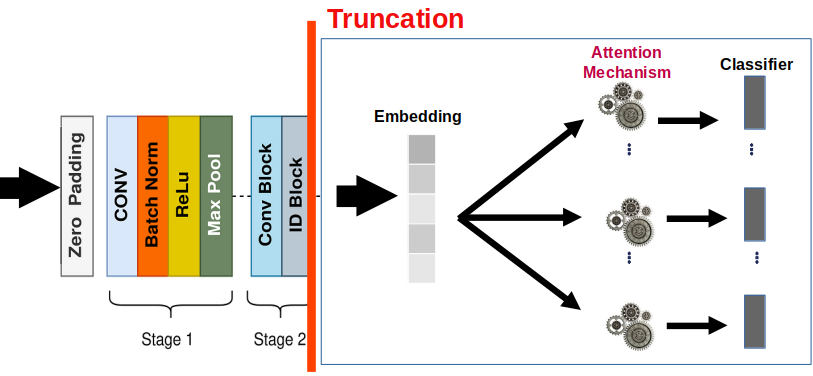}
    \caption{RTM: truncated ResNet $\rightarrow$ per-task attention $\rightarrow$ multi-task classifiers.}
    \label{fig:figure7}
\end{figure}

\paragraph{General idea.}
The core idea is to implement a truncated ResNet-50 backbone without \texttt{avgpool} to keep feature maps with $H\times W>1$, then apply per-task attention over the \emph{spatial tokens} (flattened $H\times W$) before task-specific classifiers (one head per task). See \hyperref[fig:figure7]{Fig.~\ref{fig:figure7}} for an illustration. Size and parameters: Truncated part $\approx 23.5\,\mathrm{M}$; $\sim 0.1\,\mathrm{M}$ per attention head; $4.1\,\mathrm{k}$ to $16.4\,\mathrm{k}$ per classifier (depending on the number of classes). For $12$ tasks, total $\approx 23.6\,\mathrm{M}$.

\paragraph{Pipeline.}
For an image $x\!\in\!\mathbb{R}^{3\times H\times W}$:
\begin{enumerate}
  \item \textit{Truncated encoder.} Remove \texttt{avgpool} and \texttt{fc}, then cut the block list after \texttt{truncate\_after\_layer} (e.g., \texttt{[conv1, bn1, relu, maxpool, layer1, ...]}) to obtain $F\in\mathbb{R}^{B\times C\times H'\times W'}$.

  \item \textit{Spatial tokens.} $T=\mathrm{flatten}(F) \in \mathbb{R}^{B\times (H'W')\times C}$.
  \item \textit{Per-task attention} (\texttt{TaskAttentionHead}). For each task $\tau$:
  \[
    \tilde T = T W_{\text{proj}} \in \mathbb{R}^{B\times HW\times d},\quad
    q_\tau \in \mathbb{R}^{1\times 1\times d},
  \]
  \[
    \alpha_\tau = \mathrm{softmax}\!\big(\tfrac{q_\tau \tilde T^\top}{\sqrt{d}}\big)\in\mathbb{R}^{B\times 1\times HW},\quad
    h_\tau = (\alpha_\tau \tilde T)W_{\text{out}}\in\mathbb{R}^{B\times C}.
  \]
  \item \textit{Classification head.} $\hat y_\tau = W_\tau h_\tau + b_\tau \in \mathbb{R}^{|\mathcal{Y}_\tau|}$.
\end{enumerate}

\paragraph{Loss \& weighting.}
Multi-task loss (missing labels encoded as $-1$ and ignored):
\[
\mathcal{L}=\sum_{\tau\in\mathcal{T}}\mathcal{L}_\tau,\qquad
\mathcal{L}_\tau\in\{\text{weighted CE},\ \text{Focal}(\gamma)\}.
\]
Class weights are computed \emph{per task} on the training index
(\texttt{fast\_class\_weights}, modes \texttt{hard}/\texttt{soft}/\texttt{focal}, \texttt{cap}).

\subsubsection{RTM variants}
\label{subsec:rtmg-arch}

\textbf{RTM without attention.} An ablation disables attention (\texttt{--no\_attention}) and replaces aggregation with per-task GAP + Linear. Unchanged pipeline: truncation $\rightarrow$ GAP $\rightarrow$ classifiers. Used to quantify the actual contribution of attention. Ablation detail: $h = \mathrm{GAP}(F)\in\mathbb{R}^{B\times C}$, then $\hat y_\tau = W_\tau h + b_\tau$.

\begin{figure}[h]
    \centering
    \includegraphics[width=\columnwidth]{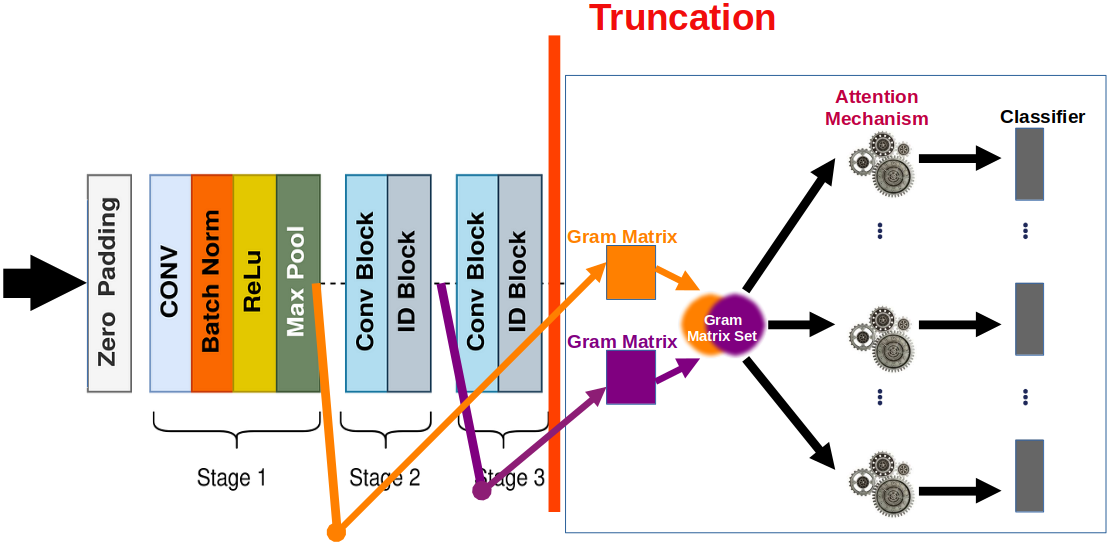}
    \caption{RTMG: truncated ResNet $\rightarrow$ Gram (style) $\rightarrow$ (attention) $\rightarrow$ multi-task heads.}
    \label{fig:figure8}
\end{figure}

\textbf{RTMG (RTM + Gram).} Adds a \emph{global Gram} step on the truncated ResNet feature maps to capture style/texture cues, followed by aggregation (with or without attention) before the multi-task heads (see \hyperref[fig:figure8]{Fig.~\ref{fig:figure8}}). Useful to test the value of an explicit style signal.

\smallskip
\noindent Full details, hyperparameters, and code:
\href{https://github.com/roadview-project/Enhanced-Style-Based-Neural-Architectures-for-Real-Time-Weather-Classification-main}{\texttt{GitHub repository}}.

\subsection{PMG: Patch-Gram + attention mechanisms with task-conditioned pooling}
\label{subsec:pmg-arch}

\begin{figure}[h]
    \centering
    \includegraphics[width=\columnwidth]{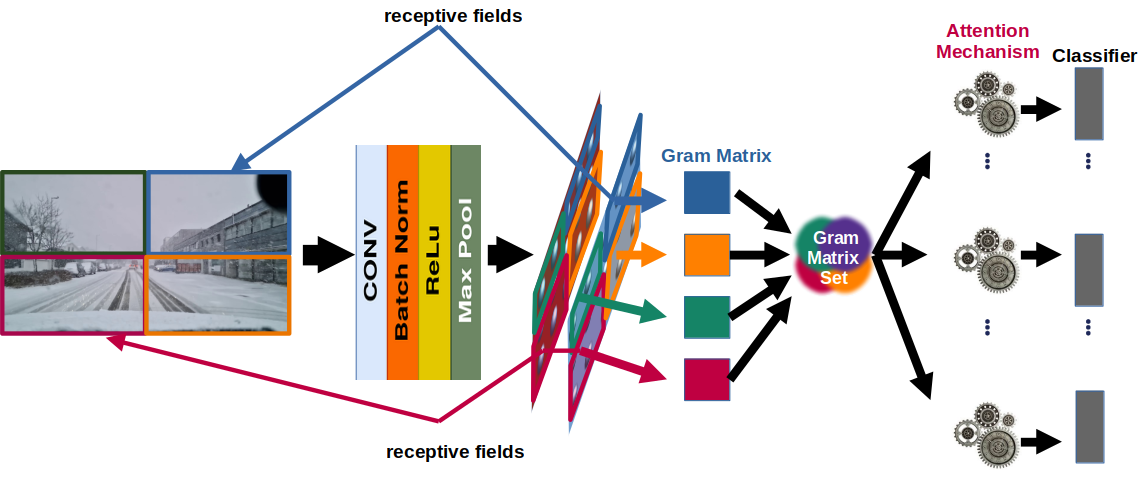}
    \caption{PMG: local Gram (style tokens) + attention mechanisms + task-conditioned pooling.}
    \label{fig:figure10}
\end{figure}

\paragraph{General idea.}
Here, we use \emph{PatchGAN} in a novel way. \emph{PMG} extracts convolutional feature maps, projects the channels, then computes and vectorizes \emph{local Gram matrices} (one per non-overlapping patch) to form \emph{style/texture tokens}. The disjoint receptive fields of PatchGAN ensure independent local Grams. These tokens are refined by a \emph{Transformer encoder} and aggregated \emph{per task} via conditioned attention pooling (a learned query per task), before multi-task \emph{linear heads} (Fig.~\ref{fig:figure10}). The total parameter count is approximately $2.43\,\mathrm{M}$ ($\sim 43\,\mathrm{k}$ for embedding extraction, $\sim 594\,\mathrm{k}$ for attention/aggregation, and $\sim 400$–$1{,}300$ per classifier).

\paragraph{Pipeline.}
For $x\!\in\!\mathbb{R}^{3\times H\times W}$:
\begin{enumerate}
    \item \emph{Conv encoder.} $F = \mathrm{FE}(x)\in\mathbb{R}^{C\times H'\times W'}$ (\texttt{feature\_extractor}, stride $4$, \texttt{BatchNorm}).
    \item \emph{Channel projection.} $F'=\mathrm{Conv}_{1\times1}(F)\in\mathbb{R}^{C_r\times H'\times W'}$, where $C_r=\texttt{gram\_channels}$.
    \item \emph{Non-overlapping patches.} Choose a spatial division $d=\texttt{patch\_div}$ to get $N_p=d^2$ patches; each patch $X_i\in\mathbb{R}^{C_r\times A}$ (with $A$ pixels after unfolding).
    \item \emph{Local Gram per patch.}
    \[
    G_i=\frac{1}{A}\,X_i X_i^\top \in \mathbb{R}^{C_r\times C_r}, \qquad
    t_i=\mathrm{vec}(G_i)\in\mathbb{R}^{C_r^2}.
    \]
    \item \emph{Token projection.} $T=[t_i]_{i=1}^{N_p}\in\mathbb{R}^{N_p\times C_r^2}$, then $T'=\mathrm{Linear}(T)\in\mathbb{R}^{N_p\times d_{\!model}}$.
    \item \emph{attention mechanisms refinement.} $\tilde T=\mathrm{attention mechanisms}(T')\in\mathbb{R}^{N_p\times d_{\!model}}$ (\texttt{attention mechanisms\_layers}, \texttt{attention mechanisms\_heads}).
    \item \emph{Task-conditioned pooling.}
    For each task $\tau$, learn a query $q_\tau\in\mathbb{R}^{d_{\!model}}$ and compute
    \[
      \alpha_{\tau,i}=\mathrm{softmax}_i\big(\langle q_\tau,\tilde t_i\rangle\big),\qquad
      z_\tau=\sum_{i=1}^{N_p}\alpha_{\tau,i}\,\tilde t_i\in\mathbb{R}^{d_{\!model}}.
    \]
    \item \emph{Classification heads.} $\hat y_\tau=W_\tau z_\tau+b_\tau$ (one linear head per task).
\end{enumerate}

\paragraph{Loss.}
Sum of per-task losses (missing labels ignored):
\[
\mathcal{L}=\sum_{\tau\in\mathcal{T}} \mathcal{L}_\tau,\quad
\mathcal{L}_\tau \in \{\text{weighted CE},\;\text{Focal}(\gamma)\}.
\]
Class weights are computed on the training split (\texttt{median} mode, cap \texttt{class\_weight\_cap}) to mitigate imbalance.

\subsubsection{PMG variants}
\label{subsec:pm-arch}

\begin{figure}[h]
    \centering
    \includegraphics[width=\columnwidth]{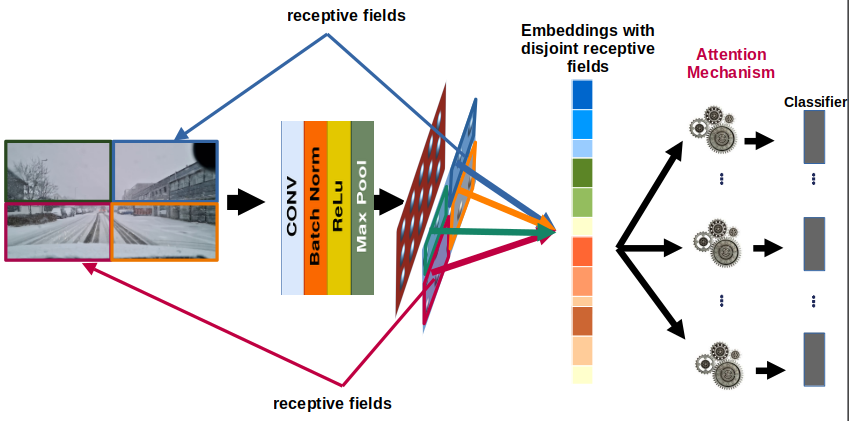}
    \caption{PM: PatchGAN trunk $\rightarrow$  per-task spatial attention $\rightarrow$ aggregation $\rightarrow$ per-task outputs.}
    \label{fig:figure9}
\end{figure}

\textbf{PM (without Gram \hyperref[fig:figure9]{Fig.~\ref{fig:figure9}}).} Variant obtained by removing PMG’s local Gram component. The \emph{PatchGAN} trunk extracts dense maps; each task applies a lightweight \emph{spatial attention} head before classification.

\textbf{PM without attention.} Ablation where the attention map is replaced by uniform pooling (GAP), to isolate the contribution of spatial weighting.

\smallskip
\noindent\emph{Full details, diagrams, and training scripts}: 
\href{https://github.com/roadview-project/Enhanced-Style-Based-Neural-Architectures-for-Real-Time-Weather-Classification-main}{\texttt{GitHub repo}}.

\subsection{Conclusion}
We designed the whole family around a unified architecture that explicitly emphasizes style-related features while preserving per-task independence and specialization. This modular design makes it easy to add or remove attributes, ensures fair comparisons across variants, and facilitates transfer, and it provides a practical testbed for studying the role of style in weather attribute prediction. Section~\ref{sec:results} presents a detailed comparative study of these architectures.

\section{Hyperparameter Optimization, Training  and Real-Time execution}

\subsection{Hyperparameter search with evolutionary algorithms}
\label{subsec:evo-hpo}

\paragraph{Summary by model and optimized hyperparameters.}
\begin{table}[!h]
\centering
\caption{Summary of models and hyperparameters optimized by evolutionary search.}
\label{tab:hyperparams}
\small
\begin{tabularx}{0.98\textwidth}{@{} l X X @{}}
\toprule
\textbf{Model} & \textbf{Summary (idea and training)} & \textbf{Optimized hyperparameters (DEAP)} \\
\midrule
\textbf{RTM} 
& Truncated ResNet-50 (no \texttt{avgpool/fc}) with lightweight \emph{per-task} attention (GAP+MLP ablation). Optimized under SGD; $K$-fold F1 as the objective. 
& \emph{truncate\_layer}, \emph{batch\_size}, \emph{lr}, head depth/width (\emph{hidden\_dims}, \emph{num\_layers}); \emph{use\_attention} (if \texttt{mode=evolve}); encoder freezing (training option). \\
\addlinespace[0.25em]
\textbf{RTMG} 
& RTM augmented with \emph{global Gram} over intermediate channels (style correlations); same training protocol (SGD, $K$-fold). 
& All RTM HPOs \emph{plus} \emph{gram\_matrix\_size}; \emph{use\_attention} (if \texttt{attention\_mode=evolve}). \\
\addlinespace[0.25em]
\textbf{PM} 
& Multi-task PatchGAN (implicit non-overlapping patches) with \emph{TaskHeadImproved} (optional SE, spatial softmax with temperature $\tau$); $K$-fold F1 objective. 
& \emph{batch\_size}, \emph{lr}, \emph{patch\_size}. \textit{Attention parameters not evolved but configurable via CLI}: \emph{attn\_tau}, \emph{attn\_use\_se}, \emph{attn\_softmax\_spatial}, \emph{attn\_tv\_lambda}. \\
\addlinespace[0.25em]
\textbf{PMG} 
& PatchGAN + \emph{local Grams} per implicit patch $\rightarrow$ token projection $\rightarrow$ (optional) lightweight attention mechanisms + task-conditioned pooling. Optimized under AdamW + Cosine; AMP/accumulation available. 
& \emph{batch\_size}, \emph{lr}, \emph{weight\_decay}, \emph{patch\_size}, \emph{patch\_div}, \emph{ndf}, \emph{gram\_channels}, \emph{d\_model}, \emph{attention mechanisms\_layers}, \emph{attention mechanisms\_heads}, \emph{use\_token\_attention}, \emph{use\_channel\_attention}, \emph{use\_focal}, \emph{focal\_gamma}. \\
\bottomrule
\end{tabularx}
\end{table}

We optimize the hyperparameters of our multi-task models using an evolutionary search (\texttt{DEAP}) coupled with $K$-fold cross-validation. The objective (\emph{fitness}) to maximize is the \emph{global mean F1}: a class-weighted mean within each task, then an average across tasks, aggregated over the $K$ folds.

\paragraph{Evaluation protocol.}
For a given hyperparameter set, we apply:
\begin{enumerate}
  \item a $K$-fold split of the training set (RTM: $K=5$; PMG: $K\in\{2,\dots\}$ depending on the script);
  \item training on $K{-}1$ folds and evaluation on the held-out fold;
  \item computation, for each task, of standard metrics (accuracy, precision, recall, class-weighted F1);
  \item aggregation of these F1 scores into a single number, used as the evolutionary \emph{fitness}.
\end{enumerate}

\paragraph{Search space.}
We explore only hyperparameters that significantly affect the model’s ability to capture style and multi-task structure.

\emph{RTM / RTMG (truncated ResNet + per-task attention).}
The encoder is a truncated ResNet-50 (without \texttt{avgpool}/\texttt{fc}), optimized with \textbf{SGD} (momentum 0.9). Main genes:
\begin{itemize}
  \item \texttt{truncate\_layer}: truncation depth (ResNet block where the network is cut);
  \item \texttt{batch\_size} and \texttt{lr}: batch size and learning rate;
  \item \texttt{hidden\_dims}, \texttt{num\_layers}: width and depth of per-task MLP heads;
  \item optional \texttt{use\_attention}: presence/absence of a per-task attention block;
  \item for RTMG, an additional parameter controls the dimension of the global Gram (\texttt{gram\_channels}).
\end{itemize}

\emph{PM / PMG (PatchGAN + local Gram + per-task pooling).}
PatchGAN models are optimized with \textbf{AdamW} and a \textbf{CosineAnnealingLR} scheduler. We vary:
\begin{itemize}
  \item \texttt{batch\_size}, \texttt{lr}, \texttt{weight\_decay};
  \item patch granularity and PatchGAN kernel/stride (\texttt{patch\_size}, \texttt{patch\_div}, \texttt{ndf});
  \item for PMG: dimensions of local Gram and tokens (\texttt{gram\_channels}, \texttt{d\_model});
  \item the lightweight attention mechanisms structure (\texttt{attention mechanisms\_layers}, \texttt{attention mechanisms\_heads});
  \item binary regularization flags (\texttt{use\_token\_attention}, \texttt{use\_channel\_attention}, \texttt{use\_focal}) and \texttt{focal\_gamma}.
\end{itemize}

Finer attention parameters (e.g., spatial softmax temperature) are fixed manually after a few preliminary trials to keep the search space reasonable.

\paragraph{Losses and class imbalance.}
For each task, we use class-weighted CrossEntropy. Weights are computed on the training indices of each fold to compensate strong asymmetries (e.g., \emph{Road Spray}, \emph{Water/Snow on Windshield}). On PMG, we additionally offer an optional Focal Loss, whose parameter $\gamma$ is also optimized by evolutionary search; in the Results section, unless stated otherwise, we report the best configurations found (weighted CrossEntropy or Focal).

\paragraph{Why evolutionary search.}
The style$\rightarrow$weather coupling induces non-linear interactions between structural hyperparameters (backbone truncation, patch granularity, local Gram width, head depth/width, attention activations) and the loss scheme (class weighting, focal). Evolutionary operators explore this mixed space (discrete/continuous/boolean) with constraints and structural ablations more effectively than standard Bayesian search, which would require costly surrogates. $K$-fold normalization stabilizes fitness in a multi-task setting (missing labels, imbalance); checkpointing ensures reproducibility and resilience.

\paragraph{Backbones and weights.}
Unless stated otherwise, ResNet-50 backbones use self-supervised MoCo-v3 weights (1{,}000 epochs)~\cite{chen2021empiricalstudytrainingselfsupervised}. Repository: \href{https://github.com/facebookresearch/moco-v3?tab=readme-ov-file}{\texttt{github.com/facebookresearch/moco-v3}}.

\subsection{Real-Time Execution on an Embedded Platform}
\label{real-time}

We evaluate continuous inference on a \emph{Raspberry Pi 5} equipped with a Broadcom BCM2712 CPU (4\(\times\) Cortex-A76 at 2.4\,GHz), a VideoCore VII GPU, LPDDR4X-4267 memory, microSD storage (256\,GB, SDR104 mode), and a 5\,V/5\,A USB-C power supply. Available interfaces include 2\(\times\)USB 3.0, 2\(\times\)USB 2.0, Gigabit Ethernet (PoE+ via HAT), Wi-Fi 802.11ac/BLE, dual HDMI 4Kp60, and a PCIe 2.0\(\times\)1 link for high-throughput peripherals. These characteristics provide a compute and I/O budget compatible with real-time inference from a single RGB camera.

\paragraph{Protocol.}
RGB video stream at 1280\(\times\)720@30\,fps, frame-wise processing without batching, 12 output heads active (multi-task), FP32 quantization (no specific NEON acceleration), measured over 2\,min with a 10\,s warm-up.

\paragraph{Throughputs obtained (mean \(\pm\) std).}
PM (2.8\,M parameters): \(30.3 \pm 1.2\) fps \quad;\quad
PMG (2.4\,M): \(25.1 \pm 1.0\) fps \quad;\quad
RTM (24\,M): \(18.4 \pm 0.9\) fps (12 tasks).
By disabling 8 of the 12 heads (same encoder), RTM reaches \(24.7 \pm 1.1\) fps.

\paragraph{Modularity and optimization.}
Thanks to the factorized architecture (one attention mechanism and one classifier per task), it is possible to:
(i) disable unneeded heads to increase throughput and reduce memory footprint;
(ii) lower the resolution to 960\(\times\)540 or 640\(\times\)360 for additional gains;
(iii) enable dynamic quantization (INT8/FP16) and an asynchronous I/O\(\rightarrow\)preprocessing\(\rightarrow\)inference pipeline.

\paragraph{Resources.}
An example embedded implementation (scripts, \texttt{systemd} service, task profiles, and video demo) is available on our GitHub repository.\footnote{\href{https://github.com/Hamedkiri/Embedded_system_rasberry}{https://github.com/Hamedkiri/Embedded\_system\_rasberry}}

\section{Results and Experiments}
\label{sec:results}

We evaluate all architectures presented in \S\ref{sec:models} on our test set,\footnote{\url{https://github.com/Hamedkiri/Weather_MultiTask_Datasets}} described in \S\ref{sec:data}. \hyperref[tab:f1_all_models_merged]{Table~\ref{tab:f1_all_models_merged}} reports F1 scores per task and the overall mean.

Four findings emerge: (i) truncating ResNet-50 yields a clear gain, with a mean F1 of \(\approx 0.98\) versus \(\approx 0.90\) for the non-truncated ResNet-50; (ii) on PatchGAN, per-task spatial attention consistently improves performance (+2 to +3 F1 points); (iii) on truncated-ResNet variants, adding attention has a limited effect under this internal protocol, but its value will be re-assessed for out-of-distribution generalization (external tests) and via t\mbox{-}SNE / Grad\mbox{-}CAM analyses; (iv) adding Gram matrices degrades performance for truncated-ResNet models but improves it for PatchGAN architectures. The latter is explained by the spatial coherence loss induced by a global Gram (as already noted by Gatys~\cite{gatys2016image}), whereas local, per-patch Gram in PatchGAN partly preserves the spatial structure of style.

Finally, PMG and RTM show similar performance on this internal set; evaluations on external data will further distinguish their generalization and robustness.

\begin{table}[!t]
\centering
\caption{F1 scores per task on the test set (25{,}000 images).
Comparison of truncated ResNet models (RTM, RTMG - 25 epochs) and PatchGAN (PM, PMG - 100 epochs). Best values per row are in bold.
RTM achieves the best overall mean, with performance close to PMG. External datasets will refine the comparison in terms of generalization and robustness.}
\label{tab:f1_all_models_merged}
\footnotesize
\renewcommand{\arraystretch}{1.15}
\setlength{\tabcolsep}{4pt}
\begin{tabular}{
  @{}
  p{0.27\linewidth}
  >{\centering\arraybackslash}p{0.095\linewidth}
  >{\centering\arraybackslash}p{0.095\linewidth}
  >{\centering\arraybackslash}p{0.095\linewidth}
  >{\centering\arraybackslash}p{0.095\linewidth}
  >{\centering\arraybackslash}p{0.095\linewidth}
  >{\centering\arraybackslash}p{0.095\linewidth}
  >{\centering\arraybackslash}p{0.095\linewidth}
  @{}
}
\toprule
\textbf{Task} & \textbf{ResNet50} & \textbf{RTM (no att.)} & \textbf{RTM (with att.)} & \textbf{RTMG} & \textbf{PM (no att.)} & \textbf{PM (with att.)} & \textbf{PMG} \\
\midrule
\multicolumn{8}{@{}l@{}}{\textit{Weather condition classification}} \\
Weather Type                    & 0.8629 & \textbf{0.9949} & 0.9944 & 0.9605 & 0.9598 & 0.9792 & 0.9877 \\
Weather Intensity               & 0.8486 & 0.9883 & \textbf{0.9908} & 0.9605 & 0.9434 & 0.9761 & 0.9845 \\
\addlinespace
\multicolumn{8}{@{}l@{}}{\textit{Visibility and sky condition}} \\
Visibility                      & 0.8542 & \textbf{0.9871} & 0.9865 & 0.9564 & 0.9382 & 0.9689 & 0.9778 \\
Sky Condition                   & 0.8686 & 0.9870 & \textbf{0.9873} & 0.9494 & 0.9491 & 0.9652 & 0.9821 \\
\addlinespace
\multicolumn{8}{@{}l@{}}{\textit{Precipitations}} \\
Precipitation Presence          & 0.8981 & \textbf{0.9956} & 0.9955 & 0.9628 & 0.9694 & 0.9823 & 0.9899 \\
Precipitation Intensity         & 0.8597 & 0.9912 & \textbf{0.9924} & 0.9504 & 0.9459 & 0.9765 & 0.9867 \\
\addlinespace
\multicolumn{8}{@{}l@{}}{\textit{Ground state}} \\
Ground Condition                & 0.8254 & \textbf{0.9750} & 0.9753 & 0.9368 & 0.9098 & 0.9601 & 0.9609 \\
\addlinespace
\multicolumn{8}{@{}l@{}}{\textit{Light and reflections}} \\
Glare / Reflections             & 0.9530 & 0.9775 & \textbf{0.9875} & 0.9845 & 0.9660 & 0.9857 & 0.9871 \\
Light Conditions                & 0.9338 & 0.9910 & \textbf{0.9944} & 0.9780 & 0.9697 & 0.9879 & 0.9931 \\
\addlinespace
\multicolumn{8}{@{}l@{}}{\textit{Road spray and windshield snow/water}} \\
Road Spray                      & 0.9578 & 0.9822 & 0.9812 & \textbf{0.9845} & 0.9586 & 0.9800 & 0.9764 \\
Water on Windshield             & 0.9207 & \textbf{0.9952} & 0.9956 & 0.9849 & 0.9771 & 0.9876 & 0.9923 \\
Snow on Windshield              & 0.9747 & \textbf{0.9984} & 0.9967 & 0.9689 & 0.9913 & 0.9957 & 0.9961 \\
\midrule
\textbf{Mean F1 score}          & 0.8994 & 0.9886 & \textbf{0.9898} & 0.9641 & 0.9565 & 0.9788 & 0.9845 \\
\bottomrule
\end{tabular}
\end{table}

\subsection{Qualitative analysis: Grad\mbox{-}CAM and t\mbox{-}SNE}

\begin{figure}[!t]
\centering
\includegraphics[width=\columnwidth]{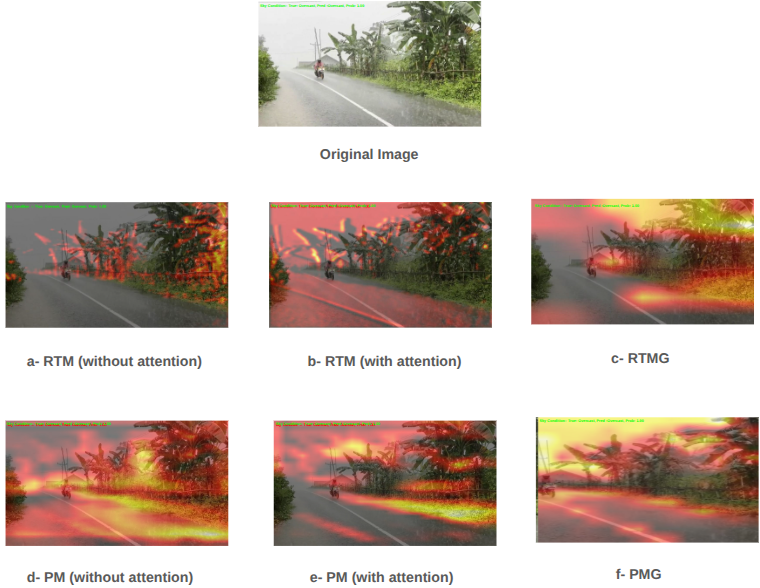}
\caption{Grad\mbox{-}CAM for the Sky Condition task.
For RTM without attention, maps are extracted at the output of the truncated ResNet-50;
for PM without attention, at the trunk output before the classifier;
for RTM and PM with attention as well as RTMG and PMG, at the output of the attention block dedicated to the task.
Overall, attention better guides focus toward relevant regions (sky, clouds).}
\label{fig:sky-condition}
\end{figure}

 \hyperref[fig:sky-condition]{Fig.~\ref{fig:sky-condition}} illustrates, using Grad\mbox{-}CAM~\cite{Selvaraju_2019}, the image regions leveraged by each model for the \emph{Sky Condition} task. Variants with attention (RTM–attn, PM–attn, RTMG, PMG) clearly concentrate activations on the sky area, indicating a relevant focus for the decision, whereas versions without attention show more diffuse maps.

\begin{figure}[h]
  \centering
  \includegraphics[width=\columnwidth]{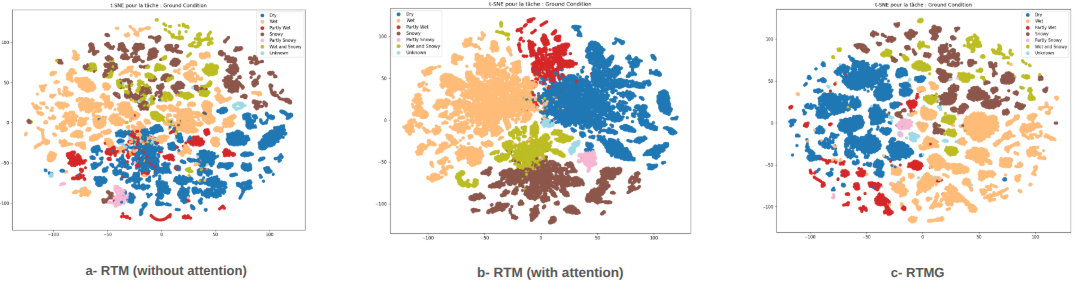}
  \caption{t\mbox{-}SNE of embeddings for \emph{Ground Condition} (50{,}388 images).
  Legend (colors as shown in the figure):
  \textcolor{blue}{Dry}, \textcolor{red!50!yellow}{Wet}, \textcolor{red}{Partly Wet},
  \textcolor{red!50!black}{Partly Snow}, \textcolor{green}{Wet and Snowy}, \textcolor{blue!60}{Unknown}.
  Subfigures: (a) RTM without attention, (b) RTM with attention, (c) RTMG.
  Attention strengthens inter-class separation (RTM with attention, RTMG), with coherent semantic proximities (\emph{Wet and Snowy} between \emph{Wet} and \emph{Snowy}).}
  \label{fig:RTM-tsne-ground}
\end{figure}

\begin{figure}[h]
  \centering
  \includegraphics[width=\columnwidth]{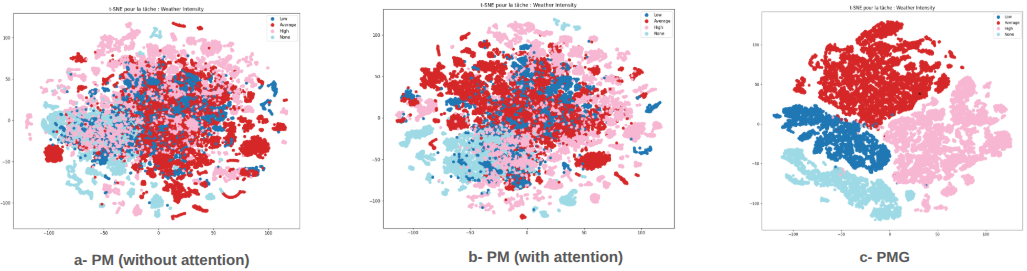}
  \caption{t\mbox{-}SNE of embeddings for \emph{Weather Intensity} (50{,}388 images).
  Legend (colors as shown in the figure):
  \textcolor{blue}{Low}, \textcolor{red}{Average}, \textcolor{magenta}{High}, \textcolor{blue!60}{None}.
  Subfigures: (a) PM without attention, (b) PM with attention, (c) PMG.
  Attention variants improve separation; PMG produces more compact clusters and sharper boundaries.}
  \label{fig:PM-Weather-Intensity}
\end{figure}

 Fig.~ \hyperref[fig:RTM-tsne-ground]{\ref{fig:RTM-tsne-ground}} and  \hyperref[fig:PM-Weather-Intensity]{\ref{fig:PM-Weather-Intensity}} present t\mbox{-}SNE~\cite{cai2022theoreticalfoundationstsnevisualizing} projections of embeddings extracted on our test set.
Overall, attention variants produce sharper inter-class separations than their non-attention counterparts. Among them, RTM and PMG yield the most compact clusters and clearest boundaries, indicating better-structured latent representations for discrimination.

This analysis using Grad\mbox{-}CAM~\cite{Selvaraju_2019} and t\mbox{-}SNE~\cite{cai2022theoreticalfoundationstsnevisualizing} shows that attention mechanisms have a real impact on the models’ ability to differentiate classes across tasks.

\subsection{Generalization capacity: \emph{zero-shot} evaluation on external datasets}
\label{generalisation-capacity}

\begin{table}[!t]
  \centering
  \footnotesize
  \setlength{\tabcolsep}{12pt}
  \renewcommand{\arraystretch}{1.2}
  \caption{\emph{Zero-shot} generalization on external datasets not seen during training (F1).
  Evaluation limited to the \emph{Weather Type} task. External labels are harmonized to \{sun, fog, rain, snow\}:
  \emph{sun} $\leftarrow$ \{Clear, Sun and Clear\},
  \emph{fog} $\leftarrow$ \{Fog, Fog and Rain, Fog and Snow\},
  \emph{rain} $\leftarrow$ \{Rain\},
  \emph{snow} $\leftarrow$ \{Snow\}.}
  \label{tab:generalisation_capacity_all}
  \begin{tabular}{@{} l c c c @{}}
    \toprule
    \textbf{Model} & \textbf{Image2Weather (F1)} & \textbf{MWI (F1)} & \textbf{Kaggle (F1)} \\
    \midrule
    RTM (truncated, no att.)  & 0.8014 & 0.8922 & 0.9660 \\
    RTM (truncated, with att.)& 0.8109 & 0.9033 & 0.9800 \\
    RTMG                      & 0.8153 & 0.8936 & 0.9730 \\
    PM (no att.)              & 0.7721 & 0.8387 & 0.9208 \\
    PM (with att.)            & 0.7567 & 0.8690 & 0.9389 \\
    PMG                       & 0.7917 & 0.8774 & 0.9527 \\
    \bottomrule
  \end{tabular}
\end{table}

\begin{figure}[!t]
  \centering
  \includegraphics[width=0.7\columnwidth]{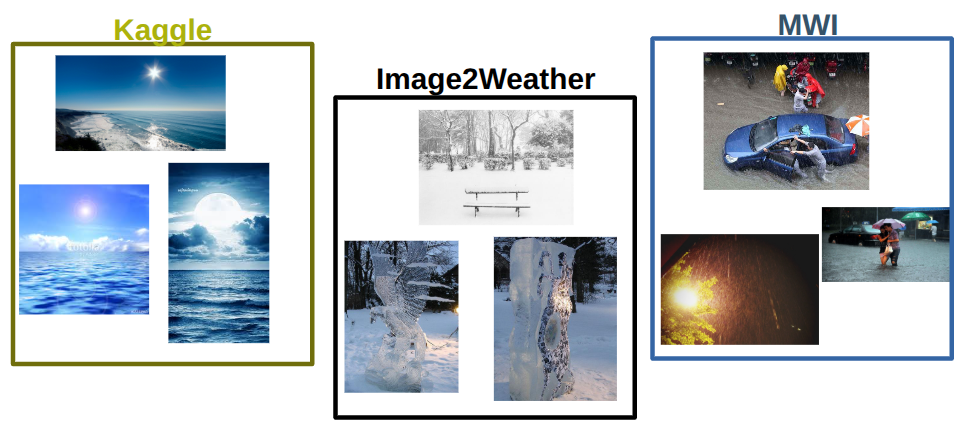}
  \caption{Example images from external datasets (unseen during training).}
  \label{fig:figure12}
\end{figure}

\begin{figure}[!t]
  \centering
  \includegraphics[width=0.7\columnwidth]{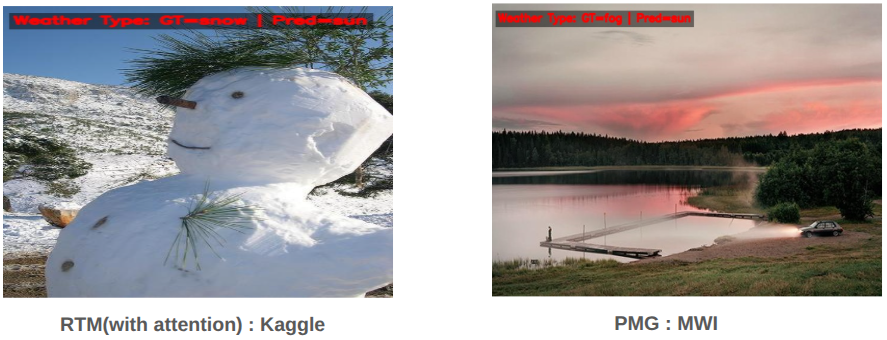}
  \caption{Examples of errors due to annotation mismatches.
  For instance, the \emph{Fog} class in MWI covers \emph{Fog}, \emph{Fog\,+\,Rain}, and \emph{Fog\,+\,Snow} in our scheme, whereas our taxonomy explicitly distinguishes precipitations.}
  \label{fig:bad-prediction-PM-RTM}
\end{figure}

We assess out-of-distribution generalization, without retraining, on three public datasets not seen during training—Image2Weather (10{,}000 images), MWI (1{,}996), and Kaggle (3{,}354)\footnote{\url{https://www.kaggle.com/datasets/vijaygiitk/multiclass-weather-dataset}}—after harmonizing labels to \{sun, fog, rain, snow\}.
The F1 scores in \hyperref[tab:generalisation_capacity_all]{Table~\ref{tab:generalisation_capacity_all}} remain high without collapse, including on images with less natural rendering (\hyperref[fig:figure12]{Fig.~\ref{fig:figure12}}).
RTM consistently achieves the best results and outperforms PMG, partly explained by a larger parameter budget (RTM \(\approx\) 23.5M vs PMG \(\approx\) 2.43M).
In general, attention improves performance when models are of comparable size, except for RTMG where the global Gram can harm spatial coherence.
\hyperref[fig:bad-prediction-PM-RTM]{Fig.~\ref{fig:bad-prediction-PM-RTM}} shows errors linked to annotation mismatches across datasets.
The evaluation script and exact label mapping are provided for reproducibility.\footnote{Code and mappings: \newline
\url{https://github.com/Hamedkiri/Heuristic_Style_Transfer_for_Real-Time_Efficient_Weather_Attribute_Detection}}

\subsection{Comparison with the literature}
\label{sec:StateOfArt}

To overcome the limitations of \emph{zero-shot} transfer (label mismatches), we evaluate the transferability of our representations by \emph{freezing the style encoder and the attention module} (Weather Type) and \emph{retraining only the classifier} on each benchmark. Training is performed on 80\,\% of the data (with 10\,\% for validation and 10\,\% for testing), in line with the protocols used in the reference works. Table~\ref{tab:ft_benchmarks_100} reports, for each dataset, two rows for our models (RTM, PMG) and one row for a state-of-the-art method.

\begin{figure}[!t]
  \centering
  \includegraphics[width=0.8\columnwidth]{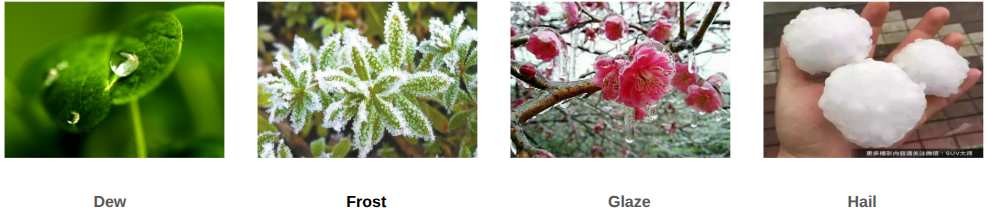}
  \caption{Example images from the external WEAPD dataset.}
  \label{fig:example-WEAPD}
\end{figure}

On MWI~\cite{zhang2015multi}, this \emph{fine-tuning} improves the performance of both PMG and RTM, and \textbf{RTM} surpasses recent methods SLM3(2024)~\cite{afxentiou2025evaluation}, ViT+Dual Attention(2023)~\cite{li2023study}. These results confirm (i) the relevance of stylistic descriptors for the \emph{Weather Type} task and (ii) their \textbf{transferability} across datasets, including under very different weather conditions, as illustrated on WEAPD (Fig.~\ref{fig:example-WEAPD}).

\begin{table}[!t]
\centering
\small
\setlength{\tabcolsep}{10pt}
\renewcommand{\arraystretch}{1.15}
\caption{Fine-tuning per benchmark (encoder and attention frozen, classifier retrained, 30 epochs). Training on 80\% of the data, validation 10\%, test 10\%. Metrics follow each reference (Accuracy or F1). Best score per block in bold.}
\label{tab:ft_benchmarks_100}
\begin{tabular}{@{} l l l @{}}
\toprule
\multicolumn{3}{@{}l@{}}{\textbf{MWI (20k images)}~\cite{zhang2015multi} \;—\; metric: Accuracy} \\
\midrule
RTM  & Acc = \textbf{0.9731} & classifier only \\
PMG  & Acc = 0.9544          & classifier only \\
\textit{SLM3 (2024)}~\cite{afxentiou2025evaluation} & Acc = 0.9001 & \textit{published value} \\
\addlinespace[4pt]
\multicolumn{3}{@{}l@{}}{\textbf{WEAPD (6{,}877 images)}~\cite{xiao2021weather} \;—\; metric: F1} \\
\midrule
RTM  & F1 = \textbf{0.9002} & classifier only \\
PMG  & F1 = 0.7874          & classifier only \\
\textit{ViT+Dual Attention (2023)}~\cite{li2023study} & F1 = 0.8769 & \textit{published value} \\
\addlinespace[4pt]
\multicolumn{3}{@{}l@{}}{\textbf{MWD (60k images)}\footnotemark \;—\; metric: F1} \\
\midrule
RTM  & F1 = \textbf{0.9903} & classifier only \\
PMG  & F1 = 0.9889          & classifier only \\
\textit{ViT+Dual Attention (2023)}~\cite{li2023study} & F1 = 0.9747 & \textit{published value} \\
\bottomrule
\end{tabular}
\end{table}
\footnotetext{\url{https://gts.ai/dataset-download/multi-class-weather-dataset}}

\paragraph{Protocol.}
Minimal label harmonization per benchmark; freezing of the backbone and the \emph{Weather Type} attention modules; classifier retraining with early stopping. Scripts, mappings, and \texttt{seeds} are available on GitHub.

\subsection{Conclusion}

Our experiments provide partial evidence that \emph{style-biased} descriptors can characterize weather conditions and their visual effects with high practical accuracy. We do not claim that style alone is sufficient to describe weather, but the results indicate that explicitly exploiting appearance cues is a promising direction.

The comparison between RTM and PMG nevertheless highlights an instructive limitation. On our internal dataset, the two models achieve very similar performance (mean F1 of 0.9845 for PMG versus 0.9898 for RTM), even though they rely on almost opposite design principles: ResNet-style architectures with residual connections progressively mix receptive fields, whereas PatchGAN-based models maintain disjoint receptive fields. The fact that their performance remains comparable reinforces the idea that \textbf{capturing stylistic information} is beneficial regardless of the encoder. However, on external datasets and for the \emph{Weather Type} task, in both \emph{zero-shot} evaluation and after classifier \emph{fine-tuning}, RTM consistently achieves the best scores, with a larger gap than observed internally. This suggests that the two architectures do not share the same inductive biases: the more capacious RTM appears to specialize more strongly on the internal distribution, while PMG, being more compact and explicitly oriented toward local style descriptors, offers a more stable trade-off between performance, computational cost, and embedded deployment.

In the absence of established reference datasets for the other tasks (intensity, ground condition, visibility, etc.), we cannot analyze these phenomena in greater depth within this work without a substantial experimental effort. We nonetheless view this as an interesting starting point: the RTM/PMG family offers two complementary regimes --- a ``large'' model targeting maximal performance (RTM) and a ``lightweight'' model designed for robustness and embedded deployment (PMG). In practice, on the \emph{Weather Type} benchmarks, PMG provides the best compromise for embedded use (small memory footprint, real-time operation), at the cost of a slight performance drop compared to larger models, a drop that can largely be mitigated by targeted \emph{fine-tuning} on the deployment context.

\section{Contributions}

We hypothesize a direct link between weather phenomena and \emph{style signatures} measurable in images (see \hyperref[fig:figure1]{Fig.~\ref{fig:figure1}}). Building on this, we systematically apply style concepts (global/local Gram, PatchGAN) to construct a unified family of lightweight, embedded-ready multi-attribute weather classifiers, validated through \emph{zero-shot} evaluations, \emph{fine-tuning}, and on an embedded platform. Our contributions are:

\begin{enumerate}
  \item \textbf{Conceptual bridge(partially): style~$\leftrightarrow$~weather.}
  We show that stylistic cues (from lower/intermediate layers) are sufficient to predict multiple attributes (\emph{Weather Type}, visibility, ground condition, etc.), by reusing and combining heuristics from style transfer.

  \item \textbf{Four style-focused architectures (Section~\href{sec:models}{\ref{sec:models}}).}
  \emph{RTM} (truncated ResNet50 + per-task attention) and \emph{RTMG} (RTM + global Gram) leverage intermediate representations;
  \emph{PM} (multi-task PatchGAN) and \emph{PMG} (PM + \emph{local} Gram) capture style in a spatially localized manner via local receptive fields.
  We introduce \emph{local} Gram matrices over PatchGAN-like “implicit patches.”
   Reproducibility and comparison: \href{https://github.com/Hamedkiri/Heuristic_Style_Transfer_for_Real-Time_Efficient_Weather_Attribute_Detection}{\texttt{Style-Based-Neural-Architectures}}.

  \item \textbf{Demonstrated transferability (Sub-sections~\href{generalisation-capacity}{\ref{generalisation-capacity}} and~\href{StateOfArt}{\ref{sec:StateOfArt}}).}
  The stylistic descriptors generalize in \emph{zero-shot} settings and after light re-training of the \emph{classifier only} on external datasets .

  \item \textbf{New multi-task implementation (12 attributes) (Section~\href{sec:models}{\ref{sec:models}}).}
  We propose a unified, modular formulation covering 12 tasks, facilitating the addition/removal of attributes and fair comparison across variants.

  \item \textbf{Real-time on constrained platforms (Sub-section~\href{real-time}{\ref{real-time}}).}
  \emph{PM} and \emph{PMG} (<\,3\,M parameters) reach real-time frame rates on Raspberry~Pi~5, with task (de)activation to adjust cost and latency. Example implementation: \href{https://github.com/Hamedkiri/Embedded_system_rasberry}{\texttt{Embedded\_system\_rasberry}}.

  \item \textbf{Large-scale open resource.}
  Release of $\sim5\times10^5$ images annotated on 12 weather attributes to support reproducibility and comparison:
  \href{https://github.com/Hamedkiri/Weather_MultiTask_Datasets}{\texttt{Weather\_MultiTask\_Datasets}}.
\end{enumerate}

These contributions establish a unified, parsimonious, and practical framework linking visual style and weather conditions, from principle through embedded deployment.

\section{Limitations, Conclusion and perspectives}
\label{sec:discussion-conclusion}

Annotating \emph{weather conditions} is intrinsically more challenging than object labeling (e.g., with \textit{bounding boxes}). In particular, \emph{intensity} (precipitation strength, visibility) is highly \emph{subjective}: the boundaries between \textit{Average} and \textit{High} are fuzzy, which introduces label noise and weakens calibration.

To reduce this uncertainty, we are building an \emph{instrumented corpus} linking road-camera images with measurements from nearby weather/airport stations. This objective \emph{ground truth} is intended to decrease annotation noise and improve the calibration of outputs.

Our experiments indicate that \emph{style-sensitive signals} (low/intermediate layers of a truncated ResNet, local receptors of the PatchGAN type, channel–channel correlations via Gram) can predict several weather attributes with high accuracy. We regard this as a \emph{partial} validation of the underlying hypothesis: scene-level semantic factors (parkas in winter, umbrellas in rain) also influence predictions and are not strictly stylistic, and we do not claim to have disentangled these components.

In the short term, we will enrich \emph{instrumented labels} (e.g., intensity in mm/h, effective visibility) for better output calibration and integrate uncertainty estimates (ECE, NLL) for ADAS use cases. In the medium term, we will profile lightweight variants on embedded GPUs/NPUs to validate \emph{in situ} \emph{real-time} operation and estimate operational indicators (e.g., water/snow quantity on the road) useful for planning and speed adaptation. Finally, we will conduct deeper studies on the role of style in attribute prediction, with the goal of an \emph{self-supervised} approach where style extraction alone would suffice to characterize weather.

\section*{Declarations}

\noindent\textbf{Funding:} This work was funded by the European Union’s Horizon Europe programme under the \href{https://roadview-project.eu/}{ROADVIEW} project (grant agreement No.~101069576).

\medskip
\noindent\textbf{Competing interests:} The authors declare no competing interests related to this article.

\medskip
\noindent\textbf{Data availability:} The \emph{Weather Multi-Task} dataset is available under a CC-BY licence at \url{https://github.com/Hamedkiri/Weather_MultiTask_Datasets}. The licences and sources of the videos, the annotation tool (images/videos/text), and the details of the train/test splits are documented in the repository.

\medskip
\noindent\textbf{Code availability:} The training and evaluation code is available at \url{https://github.com/Hamedkiri/Heuristic_Style_Transfer_for_Real-Time_Efficient_Weather_Attribute_Detection}, with textual documentation and explanatory videos.

\medskip
\noindent\textbf{Ethics approval:} No studies involving human participants or animals were conducted.

\medskip
\noindent\textbf{Consent to participate:} Not applicable.

\medskip
\noindent\textbf{Consent for publication:} Not applicable.

\medskip
\noindent\textbf{Authors' contributions:} Hamed Ouattara designed the methodology, implemented the models, and conducted the experimental analysis. \fnm{Pierre} \sur{Duthon} and Frédéric Bernardin contributed to the dataset design and the annotation protocol. All authors interpreted the results, critically revised the manuscript, and approved the final version.

\bmhead{Acknowledgements}

The authors thank the partners of the ROADVIEW consortium for their fruitful exchanges. They also thank Cerema, the Pascal Institute and the University of Clermont-Auvergne for their support.

\FloatBarrier

\end{document}